\documentclass{article}
\usepackage{ijcai23}

\usepackage{times}
\usepackage{soul}
\usepackage{url}
\usepackage[hidelinks]{hyperref}
\usepackage[utf8]{inputenc}
\usepackage[small]{caption}
\usepackage{graphicx}
\usepackage{amsmath}
\usepackage{amsthm}
\usepackage{algorithm}
\usepackage{booktabs}
\usepackage{epsfig}
\usepackage{tabu}
\usepackage{tabularray}
\usepackage[accsupp]{axessibility}
\usepackage{algorithmic}
\usepackage{amssymb}
\usepackage[switch]{lineno}
\usepackage{bbding}
\usepackage{subfig}
\usepackage{sidecap}
\usepackage{cuted}
\usepackage{multirow}
\usepackage{multicol}
\usepackage{bbm}
\usepackage{array}
\usepackage{booktabs}

\newcommand{\PreserveBackslash}[1]{\let\temp=\\#1\let\\=\temp}
\newcolumntype{C}[1]{>{\PreserveBackslash\centering}p{#1}}
\newcolumntype{R}[1]{>{\PreserveBackslash\raggedleft}p{#1}}
\newcolumntype{L}[1]{>{\PreserveBackslash\raggedright}p{#1}}

\title{A Solution to Co-occurrence Bias: Attributes Disentanglement via Mutual Information Minimization for Pedestrian Attribute Recognition}

\author{
Yibo Zhou$^1$\and
Hai-Miao Hu$^{1,2}$\and
Jinzuo Yu$^1$\and
Zhenbo Xu$^2$\and
Weiqing Lu$^1$\And
Yuran Cao$^1$
\affiliations
$^{1}$State key laboratory of virtual reality technology and systems, Beihang University\\
$^{2}$Hangzhou Innovation Institute, Beihang University\\
\emails
\{ybzhou,  hu, 17377133, 18241004\}@buaa.edu.cn, 
xuzhenbo@mail.ustc.edu.cn,
574168985@qq.com
}

\begin{document}

\maketitle

\begin{abstract}
Recent studies on pedestrian attribute recognition progress with either explicit or implicit modeling of the co-occurrence among attributes. Considering that this known a prior is highly variable and unforeseeable regarding the specific scenarios, we show that current methods can actually suffer in generalizing such fitted attributes interdependencies onto scenes or identities off the dataset distribution, resulting in the underlined bias of attributes co-occurrence. To render models robust in realistic scenes, we propose the attributes-disentangled feature learning to ensure the recognition of an attribute not inferring on the existence of others, and which is sequentially formulated as a problem of mutual information minimization. Rooting from it, practical strategies are devised to efficiently decouple attributes, which substantially improve the baseline and establish state-of-the-art performance on realistic datasets like PETAzs and RAPzs.
\end{abstract}

\section{Introduction}

Pedestrian attribute recognition (PAR), as a key component of the pedestrian analysis stemming from development of the ubiquitous video surveillance, targets to determine the soft-biometrics of local or semantic attributes of a person given its captured main-body image. To date, researches investigate PAR basically along the analogous routine of discriminative deep multi-label classification. They can be basically abstracted as emphasizing on better attribute localization to mitigate the accuracy drop from inferring on irrelevant area~\cite{liu2018localization,DBLP:conf/aaai/JiaGHCH22}, or involving additional supervision or information, like pose keypoints~\cite{liu2018localization} and pedestrian video clip~\cite{chen2019temporal,2020Pedestrian}, to guide PAR under explicit assumptions regarding body topological structure or temporal context, etc..

For work striving to enhance the attribute localization, ~\cite{fabbri2017generative,2017Learning} split the body image vertically, by a manually defined fixed strategy, into three parts and feeds each part into an individual network for feature extracting. As a further step,~\cite{liu2017hydraplus} proposed the HydraPlus-Net, learning to locate attributes of different scale with a multi-directional attention modules. Instead of learning to generate the attention map,~\cite{liu2018localization} produces attribute-specific features by means of the class activation map (CAM)~\cite{zhou2016learning}, which is initially designed to build a generic localizable deep representation. However, same technique is applied in~\cite{jia2021rethinking} to demonstrate that, even without explicitly modeling the attribute-specific area, network is still able to locate attributes precisely, implying that the fundamental task for PAR should be instead to explore better feature learning.

\begin{figure}[t]
\centering
\includegraphics[height=4.5cm,width=8.35cm]{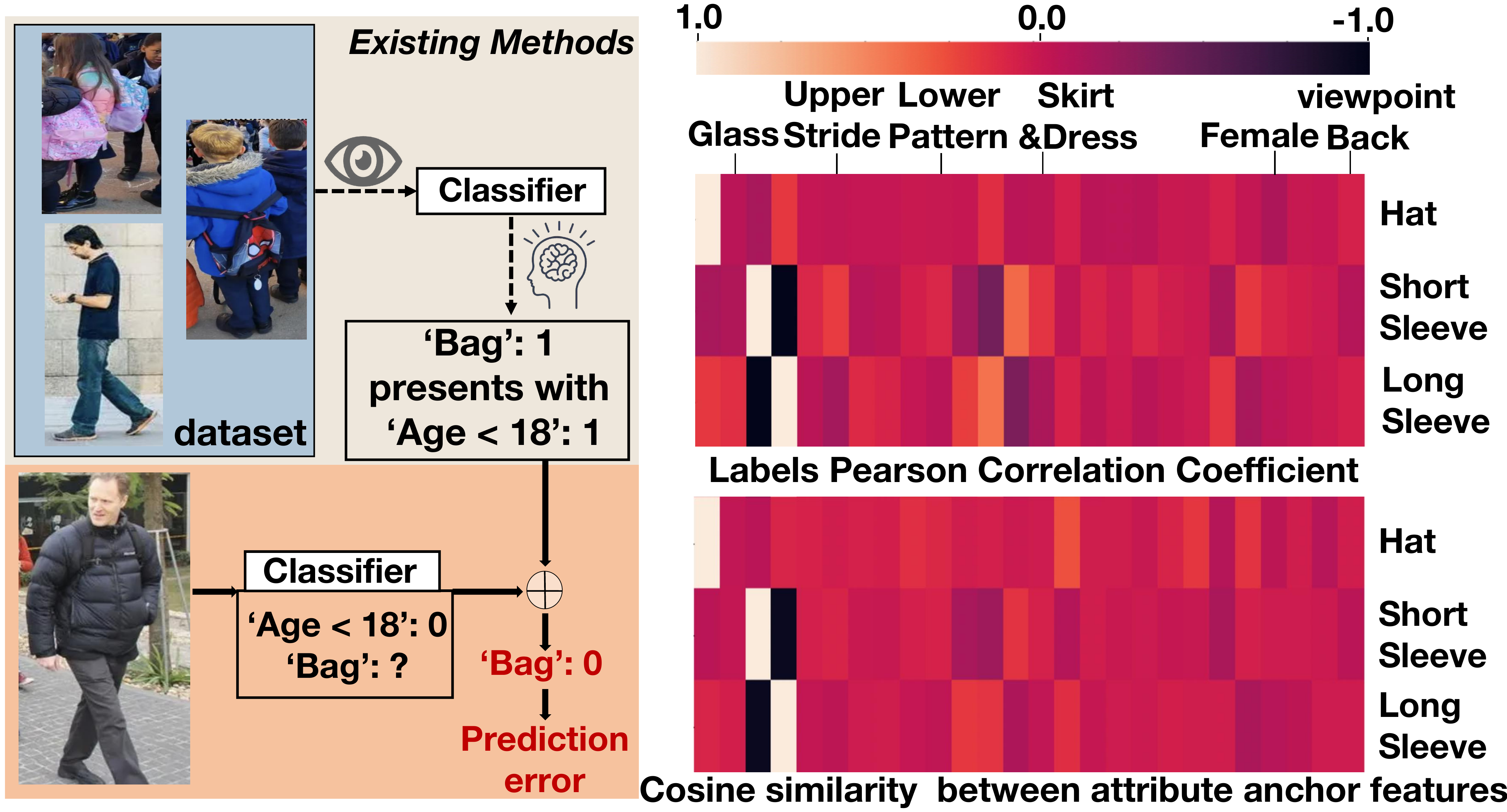}
\caption{Left: Illustration of a prediction failure resulting from biased attributes interdependency. Right: The Pearson correlation coefficient (Upper) of training set labels, and the cosine similarity between anchor features (Lower), for hat and short/long sleeve \emph{vs.} other annotated attributes in PA100k. It reveals that the network precisely captured the data selection bias that Glass \& ShortSleeve, Viewpoint-Back \& ShortSleeve and Hat \& Female, etc. are not prone to present simultaneously, and uses such biased correlations for attributes inference. The weights of the last fully-connected (FC) layer in a Resnet-50 are applied as attribute anchor features.}
\label{heatmap}
\end{figure}

Seeking for better feature learning of information exchanges, a thread of literature~\cite{2016Video,wang2017discovering,2017Multi,2020Correlation,2022Label2Label} aims to enhance the modeling of interdependencies among attributes, under the hypothesis that such correlations provide a contextual constraint complementary to visual attributes recognition. To this aim, Graph-based methods are often used to explicitly model the attributes co-occurrence and estimate the joint label probability. However, the strong variability and unpredictability exhibited by attributes co-occurrence actually cast doubts on the robustness of these methods for scenarios out of the datasets, which are typically in practical applications. Worse, given the observation in Figure.\ref{heatmap} that, such unreliable interdependency can be memorized by network, in a form of bias, even without explicit modeling of it, current methods would suffer in generalizing well onto realistic PAR, as evidenced in~\cite{jia2021rethinking}. 

Equipped with such perspective, this paper evolves in a novel spirit that we resort to infer attributes while entirely discarding their relations. Actually, it is a mechanism of PAR that accords with us human, say, we do not infer one's hat color by referring to even a single clue from its gender, instead, we look just at the hat for robustness, so should intelligent models. To embody this philosophy in stark contrast to the fragile mechanism of deep models for PAR, our attribute-disentangled learning for PAR is formalized as that the attributes speciﬁc feature learned for predicting one attribute should not use information pertinent to other attributes, i.e., the mutual information between one attribute and other attributes' specific feature should be minimized. 

To cope with the non-triviality of a direct mutual information minimization, an equivalent optimization-friendly training objective of it is deduced, which guarantees that the variation of other attributes' information results in no shift on the estimated posterior of a given attribute. Both mathematical insights and experimental evidence are provided, indicating that under this setting of the posterior-invariant learning, the proposed disentangled attribute learning can be attained, and thus tackle the issue of attribute co-occurrence bias. 

Sequentially, we also propose an efficient training strategy for the posterior-invariant learning, which enables it implemented in a manner of fast convergence. Practically, unlike most work in this field building on large networks that deteriorates the applicability, our method's plug-and-play nature makes it cater to various models, with almost no extra computational burden thereon.
Albeit lightweight, our method establishes the-state-of-art performance on various realistic datasets like the PETAzs and RAPzs~\protect\cite{jia2021rethinking}, with considerable margins over previous approaches.    

Our contribution is summarized as three folds: 

\begin{itemize} 
\item  With solid experimental evidences, we establish a novel perspective for understanding the learned attributes interdependency bias as the current bottleneck of PAR from achieving robustness. It is not covered by existing work and can serve as a mindset of future researches for rethinking PAR.
\item  We propose one direction of improvement as to infer the attributes by disregarding their correlations. From it, a lightweight prescription of information-theoretic attributes-disentangled feature learning is developed.
\item  Along with ablation studies, we present analytical experiments on various realistic PAR datasets to demonstrate our generic proposal's efficacy and a spectrum of superiorities, validating that the proposed method might serve as a foothold of robust PAR. 
\end{itemize}

\section{On the Attributes Co-occurrence Bias}
Fundamentally, although there exists certain pattern of interdependencies among attributes, such a phenomenon essentially roots from $\textbf{conditioned statistical relevance}$ rather than $\textbf{causality}$, implying that it can vary from scene to scene, individual to individual drastically. Thus, such variability and unpredictability inherent to the attributes co-occurrence make it hardly a universal known a prior to safely rely on, and could be immensely biased regarding the statistics of limited scenarios, e.g., some datasets exclusively involve indoor or outdoor scenes and constant season or whether condition, race or culture, etc.~\cite{li2016richly,liu2017hydraplus}. 

\begin{figure}[t]
\centering
\includegraphics[height=4.0cm,width=8.35cm]{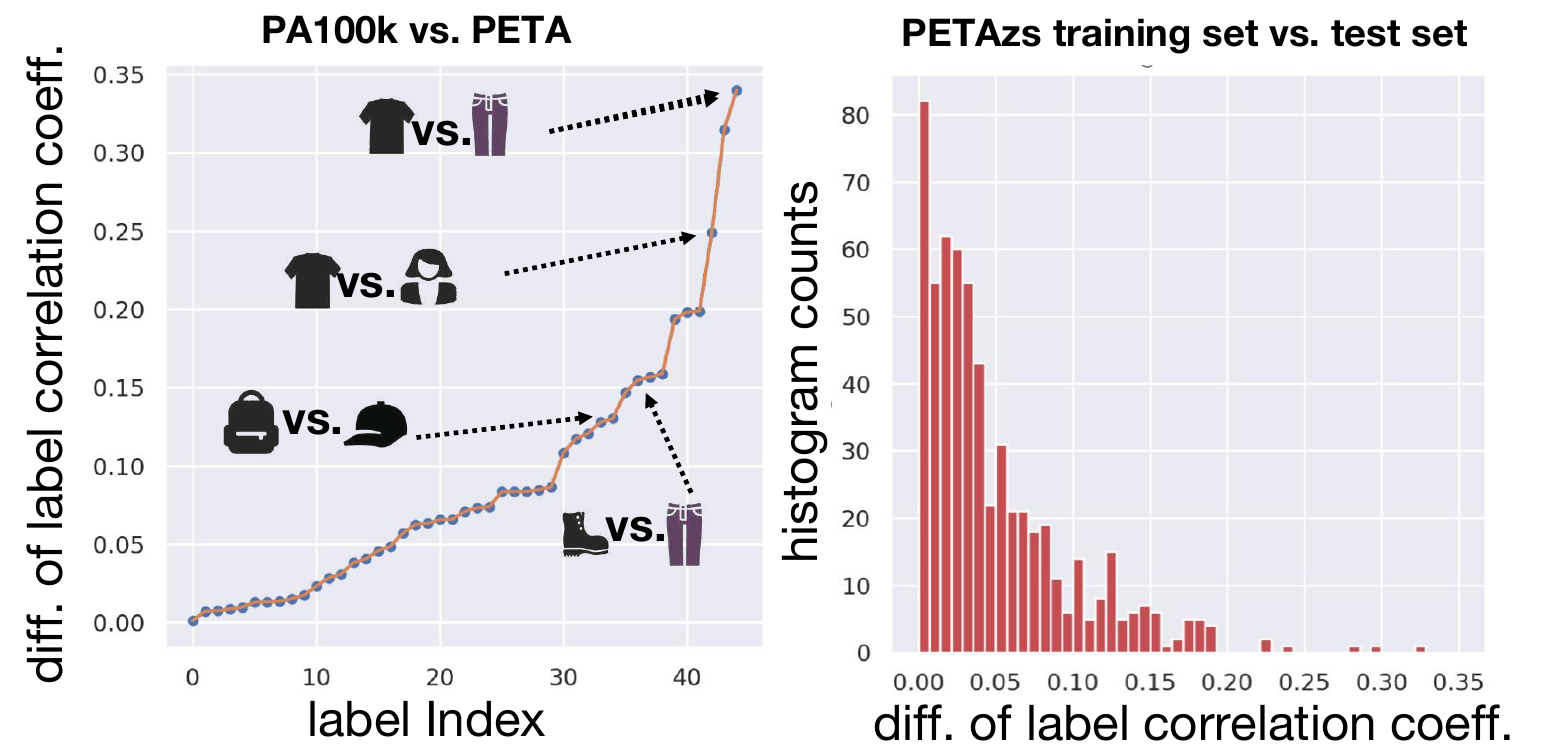}
\caption{Left: For the 10 attributes common between PA100k and PETA, the absolute difference of labels Pearson correlation coefficients computed on these two datasets. For better viewing, all values are arranged in an increasing order. Right: On the PETAzs dataset of zero-shot setting  (no overlapped identities of training set in test set), distribution of the difference between attributes Pearson correlation coefficients on training set and test set.}
\label{bias}
\end{figure}

Such a claim is supported by the observations in Figure.\ref{bias} that, the correlations among attributes are likely to fluctuate between different datasets like PA100k~\cite{liu2017hydraplus} and PETA~\cite{2014Pedestrian}, or even mutate on different groups of identities from the same scenes. For instances, ShortSleeve and Trouser lean to not co-occur in PA100k, as images in this dataset are mostly captured during summer time. While for PETA, ShortSleeve and Trouser appear simultaneously with a significantly larger ratio, resulting in the interdependency discrepancy of these two attributes. Also, even for two groups of pedestrians from a same dataset PETA, about 1/6 attributes correlations exhibit distinct characteristic (absolute difference between Pearson correlation coefficients $\ge$ 0.1). 

As a result, empirical risk dominates in the way that, existing work would have difficulty in generalizing the explicitly or implicitly leveraged attributes co-occurrence to other circumstances with different pattern of attributes interdependencies. Since an ideal dataset collection of pedestrian images, which captures global facets of the non-static myriad population distribution of attributes co-occurrence, to soften such underlined bias, can be intractable, to enhance a robust PAR, a disentangled and discriminative feature learning for each attribute can be indispensable and consequential. 

\section{Method}
\paragraph{Attributes Disentanglement by Mutual Information Minimization.}
Here, we establish the theoretical framework of our methods. Formally, it is supposed that there is a distribution $X$  characterized by all of the pedestrian images. Under certain conditions, some data points $\{ \boldsymbol x_{i} \}_{i=1}^{N}$ are sampled from $X$, with their corresponding annotations $\{ \boldsymbol a_{i} \}_{i=1}^{N}$ of some predefined attributes, to form the train set $D = \{ \boldsymbol x_{i}, \boldsymbol a_{i} \}_{i=1}^{N}$,  where $\boldsymbol a_{i} \in \{0, 1\}^C$ and $C$ is the total number of attributes. Specifically, if the latent embedding output from the feature extractor is denoted as $\boldsymbol f \in {R}^{K}$, we hope to decompose it as $\boldsymbol f=\boldsymbol f^{1}+\boldsymbol f^{2}+...+\boldsymbol f^{C}$, where$\boldsymbol f^{s} \in {R}^{K}, s = 1,2,...,C$, is the attributes-speciﬁc feature learned for predicting the attribute indicating random variable $y^{s}$. From the information theory point of view, the mutual information $\mathcal I(y^{k} ; \boldsymbol f^{s})$ measures the knowledge that could be told from the random variable $\boldsymbol f^{s}$ about the random variable $y^{k}$. Thus, to ensure the prediction of each attribute independent to the existence of others, for every $\boldsymbol f^{s}$, it should be satisfied that the mutual information $\mathcal I(y^{k} ; \boldsymbol f^{s}) = 0$, for any $k = 1,2,...,C, k \neq s$, which can be factorized as

 \begin{equation}
 \begin{aligned}
 \begin{split}
& \mathcal I(y^{k} ; \boldsymbol f^{s})= \mathcal H(y^{k}) - \mathcal H(y^{k} | \boldsymbol f^{s}) \\
=& \mathbb E_{\boldsymbol f^{s} \sim \mathcal F^{s}}[\int P(y^{k} | \boldsymbol f^{s}) \log P(y^{k} | \boldsymbol f^{s})\, dy^{k}] \\
& - \int P(y^{k}) \log P(y^{k})\, dy^{k} = 0, \label{entropy}
\end{split}
\end{aligned}
\end{equation}

\noindent where $\mathcal F^{s}$ is the marginal distribution of  $\boldsymbol f^{s}$. Noticably, if $P(y^{k} | \boldsymbol f^{s}) = P(y^{k})$,  Eq.\ref{entropy} naturally holds since $\int P(y^{k}) \log P(y^{k})\, dy^{k}$ is independent of $\boldsymbol f^{s}$. Obviously, for any $\boldsymbol f^{s}$ drawn form $\mathcal F^{s}$, if it satisfies that $P(y^{k} | \boldsymbol f^{s}) = \mathcal Q$ and $\mathcal Q$ is a constant, we have 

 \begin{equation}
 \begin{aligned}
 \begin{split}
 P(y^{k}) & = \int P( \boldsymbol f^{s}) P(y^{k} | \boldsymbol f^{s})\, d \boldsymbol f^{s}\\
 & = \int P( \boldsymbol f^{s}) \mathcal Q \, d \boldsymbol f^{s} = \mathcal Q. \label{equal}
\end{split}
\end{aligned}
\end{equation}

\noindent Eq.\ref{equal} reveals that, to satisfy $P(y^{k}) = P(y^{k} | \boldsymbol f^{s})$ is equivalent to ensure that for any $\boldsymbol f^{s}_a$ and $\boldsymbol f^{s}_b$ sampled from $\mathcal F^{s}$, it holds that $P(y^{k} |  \boldsymbol f^s = \boldsymbol f^{s}_a) = P(y^{k} | \boldsymbol f^s = \boldsymbol f^{s}_b)$. Note that 

 \begin{small}
 \begin{equation}
 \begin{aligned}
 \begin{split}
&P(y^{k} | \boldsymbol f^s = \boldsymbol f^{s}_a) - P(y^{k} |  \boldsymbol f^s = \boldsymbol f^{s}_b) \\
= &\int (P(y^{k} | \boldsymbol f^{s}_a+ \sum_{l=1, l \neq s}^C \boldsymbol f^{l}) - P(y^{k} | \boldsymbol f^{s}_b+ \sum_{l=1, l \neq s}^C \boldsymbol f^{l}) )\\
&\cdot P(\boldsymbol f^{1},\boldsymbol f^{2},..,\boldsymbol f^{s-1},\boldsymbol f^{s+1},..,\boldsymbol f^{C})\, \prod_{l=1, l \neq s}^{n} d \boldsymbol f^{l}. \nonumber
\end{split}
\end{aligned}
\end{equation}
 \end{small}

\noindent With this formulation, if the estimate of $P(y^{k} | \boldsymbol f) = P(y^{k} | \sum_{l=1}^C \boldsymbol f^{l})$ is invariant of $\boldsymbol f$'s component $\boldsymbol f^{s}$, for any $s \neq k$, then we have $\mathcal I(y^{k} ; \boldsymbol f^{s}) = 0$. Thus, a statement can be declared as follows
\begin{center}
\fbox
{\shortstack[c]{
\textbf{Goal}: attribute-disentangled learning can be enabled as \\
long as the probability estimate of $P(y^{k} | \boldsymbol f)$ varies only \\
with the $y^{k}$'s specific feature $\boldsymbol f^{k}$.
}
}
\end{center}

\begin{figure*}[t]
\centering
\subfloat[]{
\begin{minipage}[t]{0.75\linewidth} 
\centering
\includegraphics[width=15.15cm]{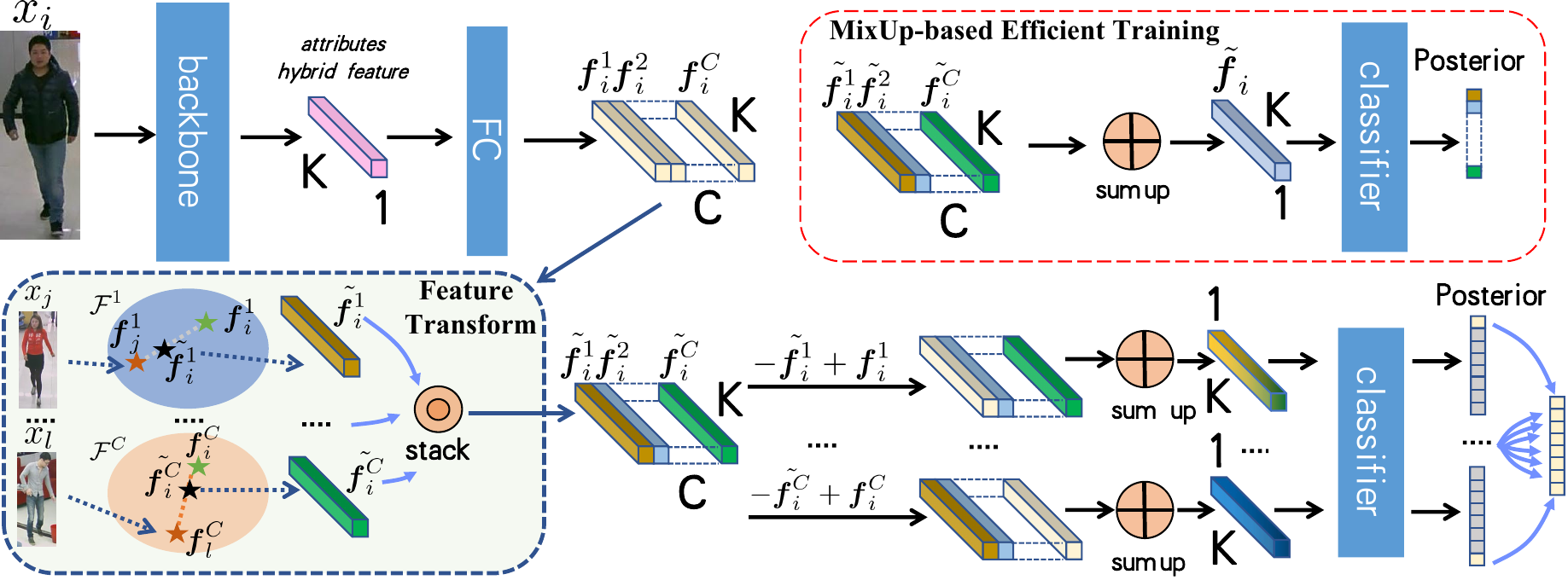}
\end{minipage} \label{framework}
}%
\subfloat[]{
\begin{minipage}[t]{0.34\linewidth}
\centering
\includegraphics[width=2.25cm]{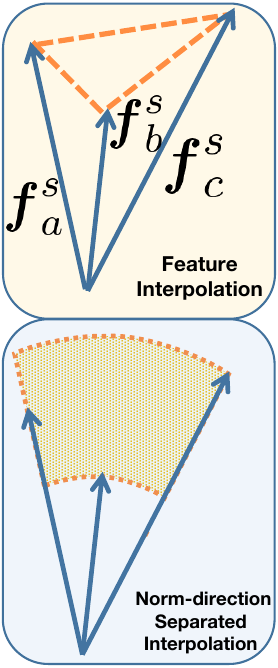}
\end{minipage} \label{holder}
}%
\caption{(a): Overall pipeline of the proposed method for attribute-disentangled feature learning. For Eq.\ref{train}, the transformed attribute-specific features are fed into another branch within the red-dashed box. (b): Comparison of direct feature interpolation (Upper) and the proposed norm-direction separated interpolation (Lower), in term of the domain exploration of feature distribution. Given three attribute-specific features of same label, by direct feature interpolation only points on the orange line are regarded as possible variations of them. Whilst for norm-direction separated interpolation, in a reliable manner, the potential variations are expanded into the whole orange-shaded area.}
\end{figure*}

\paragraph{Approach of Posterior-invariant Learning.}
In practice, for the feature of attributes hybrid, extracted from a given input $\boldsymbol x_i$, we translate it first to obtain each attribute-specific components $\boldsymbol f_i^s, s = 1,2,...,C$, by a FC layer followed with nonlinearities, and $\boldsymbol f_i $ is obtained by adding them up as $\boldsymbol f_i = \sum_{s=1}^C \boldsymbol f_i^s$. Sequentially, we randomly build $C$ mappings $\mathcal G^s_i(\cdot): \mathcal F^{s} \mapsto \mathcal F^{s}$ to transform each $\boldsymbol f_i^s$ into an arbitrary data point $\tilde{\boldsymbol f_i^s}$ within $\boldsymbol f^s$'s domain of distribution $\mathcal F^s$, and generate a new feature vector of $\tilde{\boldsymbol f_i} = \sum_{s=1}^C \tilde{\boldsymbol f^{s}_i},$ where $\tilde{\boldsymbol f^{s}_i} = \mathcal G_i^s(\boldsymbol f^{s}_i)$. We require $P(y^{s} = y_i^{s} | \tilde{\boldsymbol f_i} - \tilde{\boldsymbol f_i^s} + \boldsymbol f_i^s) = P(y^{s} = y_i^{s} | \boldsymbol f_i)$ to meet the goal of the proposed attribute-disentangled learning that $P(y^{s} | \boldsymbol f)$ should be tolerant to the variation of other attributes' specific features, and therefore to ensure no factors for decreasing the uncertainty of $y^{s}$ are conveyed by $\boldsymbol f^k, \forall k \neq s$. Under the context of supervised PAR, $P(y^{s} = y_i^{s} | \boldsymbol f_i)$ is simply the ground truth label $a_i^s$ coupled with the input $\boldsymbol x_i$, and thereby the underlined attribute-disentangled learning can be enabled by leveraging the information-theoretic regularizer of

 \begin{equation}
 \begin{aligned}
 \begin{split}
\min - &\sum_{i=1}^N \sum_{s=1}^C a_i^s \log \widehat{P}(y^{s} = y_i^{s} | \tilde{\boldsymbol f_i} - \tilde{\boldsymbol f_i^s} + \boldsymbol f_i^s), \\
&s.t. \,\,\, \tilde{\boldsymbol f_i} = \sum_{s=1}^C \tilde{\boldsymbol f_i^s}, \,\,and\,\,\,  \tilde{\boldsymbol f_i^s} = \mathcal G_i^s(\boldsymbol f^{s}_i).  \label{regular}
\end{split}
\end{aligned}
\end{equation}

\noindent $\widehat{P}$ is the estimate of $P(y^{k} | \boldsymbol f)$ given by a discriminative classifier, which is a common technique for approximating a posterior, and plugged right onto $\boldsymbol f$ with MLP operations in it. This framework is graphed in Figure.\ref{framework}.

To specify the form of $\mathcal G_i^s(\cdot)$, we opt for simplicity by adopting the convex linear combination between $\boldsymbol f_i^s$ and $\boldsymbol f_{r(i,s)}^s$, which is the extracted feature from a randomly picked training sample $\boldsymbol x_{r(i,s)}$, as a plausible form among possible variants that satisfy the closeness of a transform, i.e., the mapped feature still lies within the domain of $\mathcal F^s$. Here, $r(i,s)$ is simply a function that maps different $i \& s$ to a different sample index. It might be noticed that, with the convex combination of two attribute-specific features from training samples, points off of the training distribution can be generated. However,~\cite{2013Efficient} has shown that this linear interpolation between hidden states is an effective way of transiting between learned factors to produce new in-distribution semantics, making the combined features still informative of attributes thus reside within the $\mathcal F^s$. Specifically, 

\begin{small}
\begin{equation}
\begin{aligned}
\begin{split}
\mathcal G_i^s(\boldsymbol f^{s}_i) = \left \{
\begin{array}{ll}
    (\beta^s \boldsymbol \Vert \boldsymbol f^{s}_i \Vert + (1 - \beta^s) \boldsymbol \Vert \boldsymbol f^{s}_{r(i,s)} \Vert)   \\
    \cdot (\alpha^s \frac{ \boldsymbol f^{s}_i}{\Vert  \boldsymbol f^{s}_i \Vert} + (1 - \alpha^s) \frac{\boldsymbol f^{s}_{r(i,s)}}{ \Vert \boldsymbol f^{s}_{r(i,s)} \Vert})    & if \,\,\, a_i^s = a_{r(i,s)}^s\\
    \\
    \\
     \alpha^s \boldsymbol f^{s}_i + (1 - \alpha^s) \boldsymbol f^{s}_{r(i,s)}            & otherwise. \nonumber
\end{array}
\right.
\end{split}
\end{aligned}
\end{equation}
\end{small}

\noindent $\alpha^s$ and $\beta^s$ are independently drawn from the uniform distribution $\mathcal U(0,1)$ for each attribute. To fully explore $\mathcal F_i$ and produce more of the points with semantics potentially to be presented in the capricious test environment, for attribute-specific features with identical label, we do the linear combinations to their norms and direction vectors, respectively. The rationale is that, since feature norm has been validated to be an effective measure of data uncertainty~\cite{shalev2018out,Meng_2021_CVPR,2022Rethinking} that is prone to be fluctuated by bad image quality (often in PAR), the style semantics of attribute (color, size, shape, etc.) might be mostly encoded into feature's direction, i.e., it's normalized unit feature~\cite{2018Additive}. Hence, such a norm-direction-separated linear combination can generate features from attributes of one style but shot under various uncertainties, exploiting more of the domain of $\mathcal F_s$, exemplified in Figure.\ref{holder}.

\paragraph{Efficient MixUp-based Alternative.}
Notably, directly optimizing over Eq.\ref{regular} necessitates $C$ inferences though the classifier for each sample in every update, which can be computationally inefficient when the number of attributes grows larger. Regardless of the ease for implementing $\mathcal G(\cdot)$ as linear combination, such a design enjoys another merit as it actually supplies us with an approach to integrate the training objective of Eq.\ref{regular} into a desirable single-inference-every-instance manner of

 \begin{equation}
 \begin{aligned}
 \begin{split}
&\min - \sum_{i=1}^N\sum_{s=1}^C (\alpha^s a_i^s + (1 - \alpha^s) a_{r(i,s)}^s)\\
& \,\,\, \,\,\,\,\,\, \,\,\,\,\,\, \,\,\, \cdot \log \widehat{P}(y^{s} = \alpha^s y_i^s + (1 - \alpha^s) y_{r(i,s)}^s | \tilde{\boldsymbol f_i}),\\
&s.t. \,\,\, \tilde{\boldsymbol f_i} = \sum_{s=1}^C \tilde{\boldsymbol f_i^s}, \,\,\,\,\,\, and\,\,\,\, \tilde{\boldsymbol f_i^s}, \alpha^s, a_{r(i,s)}^s = \mathcal G_i^s(\boldsymbol f^{s}_i).
\label{train}
\end{split}
\end{aligned}
\end{equation}

\noindent Such a simplification from Eq.\ref{regular} to Eq.\ref{train} is inspired by the well-known data augmentation method MixUp~\cite{zhang2017mixup}, which shows that the linear combination of images is actually aligned with exactly the same linear combination of their ground truth in the label space. Here, we adopt an analogous idea into Eq.\ref{train} to reduce the training cost. 

Please note that Eq.\ref{train} differs from MixUp in two aspects. First, during every update, MixUp uses a single $\alpha$ and a single sample-to-interpolate for each input, purely for the sake of data augmentation. Whereas in Eq.\ref{train}, for each attribute's specific feature, we use feature interpolation of different $\alpha^s$, $\beta^s$ and $\boldsymbol f_{r(i,s)}^s$ to guarantee that the transform imposed over it is totally independent thus differentiated from others. By requiring the variation of a certain attribute's posterior aligned exclusively to the variation of its specific feature, we enforce other attribute‘s specific feature acquires no factors informative to the given attribute's posterior, exactly the same way for realizing attribute-disentangled learning by Eq.\ref{regular}. Second, instead of pixel space, the linear combination is conducted in the space of decoupled attribute-specific features. One might argue that such a strategy can bring non-trivial boost in performance by somewhat working as a data augmentation of MixUp. We admit this concern. However, we will present in the experiment section that it is actually not the case here for achieving SOTA performance as MixUp can not foster any accuracy increase over the baseline. More importantly, as will be presented in the experiment section, the equivalent Eq.\ref{regular} sufﬁces to deliver similar performance w.r.t Eq.\ref{train}, which does not optimize over the interpolated feature by its correspondingly interpolated label.

\label{method}

\begin{table*}[t]
\setlength{\belowcaptionskip}{-0.cm}
\footnotesize
\centering
\resizebox{\textwidth}{!}{
            \renewcommand{\arraystretch}{1.2}
            \begin{tabular}{cccc|cccccc|cccccc|ccc|ccc}
               \toprule
	\multicolumn{4}{c|}{\multirow{2}{*}{\textbf{Method}}}&\multicolumn{3}{c|}{\multirow{2}{*}{\textbf{Backbone}}}&\multicolumn{3}{c|}{\textbf{PA100k}}&\multicolumn{3}{c}{\textbf{RAP}}&\multicolumn{3}{c|}{\textbf{RAPzs}}&\multicolumn{3}{c}{\textbf{PETA}}&\multicolumn{3}{c}{\textbf{PETAzs}}\\	
	\cline{8-22}
	\multicolumn{4}{c|}{}&\multicolumn{3}{c|}{}&\multicolumn{1}{c}{\textbf{mA}}&\multicolumn{1}{c}{\textbf{Recall}}&\multicolumn{1}{c|}{\textbf{F1}}&\multicolumn{1}{c}{\textbf{mA}}&\multicolumn{1}{c}{\textbf{Recall}}&\multicolumn{1}{c}{\textbf{F1}}&\multicolumn{1}{c}{\textbf{mA}}&\multicolumn{1}{c}{\textbf{Recall}}&\multicolumn{1}{c|}{\textbf{F1}}&\multicolumn{1}{c}{\textbf{mA}}&\multicolumn{1}{c}{\textbf{Recall}}&\multicolumn{1}{c}{\textbf{F1}}&\multicolumn{1}{c}{\textbf{mA}}&\multicolumn{1}{c}{\textbf{Recall}}&\multicolumn{1}{c}{\textbf{F1}} \\
 	\hline
	\hline
	\multicolumn{4}{c|}{\multirow{1}{*}{Baseline ('21)}}&\multicolumn{3}{c|}{Resnet-50}&\multicolumn{1}{c}{80.38}&\multicolumn{1}{c}{87.01}&\multicolumn{1}{c|}{87.05}&\multicolumn{1}{c}{80.32}&\multicolumn{1}{c}{79.89}&\multicolumn{1}{c}{79.46}&\multicolumn{1}{c}{72.32}&\multicolumn{1}{c}{76.62}&\multicolumn{1}{c|}{76.75}&\multicolumn{1}{c}{84.42}&\multicolumn{1}{c}{85.08}&\multicolumn{1}{c}{85.97}&\multicolumn{1}{c}{71.62}&\multicolumn{1}{c}{70.33}&\multicolumn{1}{c}{71.68}\\
	\hline
	\multicolumn{4}{c|}{MsVAA ('18)}&\multicolumn{3}{c|}{Resnet-50}&\multicolumn{1}{c}{80.41}&\multicolumn{1}{c}{86.52}&\multicolumn{1}{c|}{86.80}&\multicolumn{1}{c}{78.86}&\multicolumn{1}{c}{79.15}&\multicolumn{1}{c}{79.27}&\multicolumn{1}{c}{72.04}&\multicolumn{1}{c}{75.81}&\multicolumn{1}{c|}{75.74}&\multicolumn{1}{c}{84.35}&\multicolumn{1}{c}{85.51}&\multicolumn{1}{c}{86.09}&\multicolumn{1}{c}{71.53}&\multicolumn{1}{c}{69.42}&\multicolumn{1}{c}{71.94} \\
	\multicolumn{4}{c|}{MTMS ('19)}&\multicolumn{3}{c|}{Resnet-50}&\multicolumn{1}{c}{-}&\multicolumn{1}{c}{-}&\multicolumn{1}{c|}{-}&\multicolumn{1}{c}{82.45}&\multicolumn{1}{c}{80.44}&\multicolumn{1}{c}{65.33}&\multicolumn{1}{c}{-}&\multicolumn{1}{c}{-}&\multicolumn{1}{c|}{-}&\multicolumn{1}{c}{86.23}&\multicolumn{1}{c}{87.22 }&\multicolumn{1}{c}{85.85}&\multicolumn{1}{c}{-}&\multicolumn{1}{c}{-}&\multicolumn{1}{c}{-}\\
	\multicolumn{4}{c|}{VAC ('20)}&\multicolumn{3}{c|}{Resnet-50}&\multicolumn{1}{c}{79.16}&\multicolumn{1}{c}{86.26}&\multicolumn{1}{c|}{87.59}&\multicolumn{1}{c}{80.27}&\multicolumn{1}{c}{79.77}&\multicolumn{1}{c}{78.36}&\multicolumn{1}{c}{73.70}&\multicolumn{1}{c}{76.97}&\multicolumn{1}{c|}{76.12}&\multicolumn{1}{c}{83.63}&\multicolumn{1}{c}{85.45}&\multicolumn{1}{c}{86.23}&\multicolumn{1}{c}{71.91}&\multicolumn{1}{c}{70.64}&\multicolumn{1}{c}{70.90}\\
	\multicolumn{4}{c|}{*JLAC ('20)}&\multicolumn{3}{c|}{Resnet-50}&\multicolumn{1}{c}{82.31}&\multicolumn{1}{c}{87.77}&\multicolumn{1}{c|}{87.61}&\multicolumn{1}{c}{83.69}&\multicolumn{1}{c}{82.40}&\multicolumn{1}{c}{\textbf{80.82}}&\multicolumn{1}{c}{76.38}&\multicolumn{1}{c}{79.20}&\multicolumn{1}{c|}{76.05}&\multicolumn{1}{c}{86.96}&\multicolumn{1}{c}{87.09}&\multicolumn{1}{c}{\textbf{87.45}}&\multicolumn{1}{c}{73.60}&\multicolumn{1}{c}{\textbf{72.41}}&\multicolumn{1}{c}{72.05} \\
	\multicolumn{4}{c|}{SSC ('21)}&\multicolumn{3}{c|}{Resnet-50}&\multicolumn{1}{c}{81.87}&\multicolumn{1}{c}{89.10}&\multicolumn{1}{c|}{86.87}&\multicolumn{1}{c}{82.77}&\multicolumn{1}{c}{\textbf{87.49}}&\multicolumn{1}{c}{80.43}&\multicolumn{1}{c}{-}&\multicolumn{1}{c}{-}&\multicolumn{1}{c|}{-}&\multicolumn{1}{c}{86.52}&\multicolumn{1}{c}{87.12}&\multicolumn{1}{c}{86.99}&\multicolumn{1}{c}{-}&\multicolumn{1}{c}{-}&\multicolumn{1}{c}{-}\\
	\multicolumn{4}{c|}{DAFL ('22)}&\multicolumn{3}{c|}{Resnet-50}&\multicolumn{1}{c}{83.54}&\multicolumn{1}{c}{89.19}&\multicolumn{1}{c|}{\textbf{88.90}}&\multicolumn{1}{c}{83.72}&\multicolumn{1}{c}{83.39}&\multicolumn{1}{c}{80.29}&\multicolumn{1}{c}{-}&\multicolumn{1}{c}{-}&\multicolumn{1}{c|}{-}&\multicolumn{1}{c}{87.07}&\multicolumn{1}{c}{87.03}&\multicolumn{1}{c}{86.40}&\multicolumn{1}{c}{-}&\multicolumn{1}{c}{-}&\multicolumn{1}{c}{-}\\
	\multicolumn{4}{c|}{Label2Label ('22)}&\multicolumn{3}{c|}{Resnet-50}&\multicolumn{1}{c}{82.24}&\multicolumn{1}{c}{88.57}&\multicolumn{1}{c|}{87.08}&\multicolumn{1}{c}{-}&\multicolumn{1}{c}{-}&\multicolumn{1}{c}{-}&\multicolumn{1}{c}{73.84}&\multicolumn{1}{c}{78.15}&\multicolumn{1}{c|}{77.75}&\multicolumn{1}{c}{-}&\multicolumn{1}{c}{-}&\multicolumn{1}{c}{-}&\multicolumn{1}{c}{72.13}&\multicolumn{1}{c}{71.10}&\multicolumn{1}{c}{71.74}\\
	\hline
	\multicolumn{4}{c|}{\textbf{Ours}}&\multicolumn{3}{c|}{Resnet-50}&\multicolumn{1}{c}{84.53}&\multicolumn{1}{c}{89.13}&\multicolumn{1}{c|}{87.01}&\multicolumn{1}{c}{83.26}&\multicolumn{1}{c}{83.31}&\multicolumn{1}{c}{79.88}&\multicolumn{1}{c}{76.71}&\multicolumn{1}{c}{80.18}&\multicolumn{1}{c|}{77.93}&\multicolumn{1}{c}{86.41}&\multicolumn{1}{c}{86.80}&\multicolumn{1}{c}{87.03}&\multicolumn{1}{c}{74.35}&\multicolumn{1}{c}{72.09}&\multicolumn{1}{c}{72.39} \\
	\multicolumn{4}{c|}{\textbf{*Ours}}&\multicolumn{3}{c|}{Resnet-50}&\multicolumn{1}{c}{\textbf{85.12}}&\multicolumn{1}{c}{\textbf{89.40}}&\multicolumn{1}{c|}{87.85}&\multicolumn{1}{c}{\textbf{83.88}}&\multicolumn{1}{c}{83.65}&\multicolumn{1}{c}{80.73}&\multicolumn{1}{c}{\textbf{77.49}}&\multicolumn{1}{c}{\textbf{80.42}}&\multicolumn{1}{c|}{\textbf{78.30}}&\multicolumn{1}{c}{\textbf{87.18}}&\multicolumn{1}{c}{\textbf{87.29}}&\multicolumn{1}{c}{87.41}&\multicolumn{1}{c}{\textbf{75.13}}&\multicolumn{1}{c}{71.98}&\multicolumn{1}{c}{\textbf{72.49}} \\
	\hline
	\hline
	\multicolumn{4}{c|}{\multirow{1}{*}{Baseline ('22)}}&\multicolumn{3}{c|}{ConvNeXt-base}&\multicolumn{1}{c}{82.2-}&\multicolumn{1}{c}{-}&\multicolumn{1}{c|}{88.5-}&\multicolumn{1}{c}{-}&\multicolumn{1}{c}{-}&\multicolumn{1}{c}{-}&\multicolumn{1}{c}{76.54}&\multicolumn{1}{c}{81.47}&\multicolumn{1}{c|}{80.25}&\multicolumn{1}{c}{86.1-}&\multicolumn{1}{c}{-}&\multicolumn{1}{c}{88.1-}&\multicolumn{1}{c}{75.21}&\multicolumn{1}{c}{74.43}&\multicolumn{1}{c}{75.62} \\
	\hline
	\multicolumn{4}{c|}{ALM ('19)}&\multicolumn{3}{c|}{BN-inception}&\multicolumn{1}{c}{80.68}&\multicolumn{1}{c}{88.84}&\multicolumn{1}{c|}{86.46}&\multicolumn{1}{c}{81.87}&\multicolumn{1}{c}{86.48}&\multicolumn{1}{c}{80.16}&\multicolumn{1}{c}{74.28}&\multicolumn{1}{c}{80.73}&\multicolumn{1}{c|}{76.65}&\multicolumn{1}{c}{84.24}&\multicolumn{1}{c}{85.60}&\multicolumn{1}{c}{85.41}&\multicolumn{1}{c}{73.01}&\multicolumn{1}{c}{73.69}&\multicolumn{1}{c}{71.53} \\
	\multicolumn{4}{c|}{MTA-Net ('20)}&\multicolumn{3}{c|}{Resnet-152}&\multicolumn{1}{c}{-}&\multicolumn{1}{c}{-}&\multicolumn{1}{c|}{-}&\multicolumn{1}{c}{77.62}&\multicolumn{1}{c}{78.44}&\multicolumn{1}{c}{79.07}&\multicolumn{1}{c}{-}&\multicolumn{1}{c}{-}&\multicolumn{1}{c|}{-}&\multicolumn{1}{c}{84.62}&\multicolumn{1}{c}{86.42}&\multicolumn{1}{c}{86.04}&\multicolumn{1}{c}{-}&\multicolumn{1}{c}{-}&\multicolumn{1}{c}{-} \\
	\multicolumn{4}{c|}{AR-BiFPN ('20)}&\multicolumn{3}{c|}{EfficientNet-B3}&\multicolumn{1}{c}{81.45}&\multicolumn{1}{c}{89.46}&\multicolumn{1}{c|}{87.94}&\multicolumn{1}{c}{82.37}&\multicolumn{1}{c}{\textbf{87.23}}&\multicolumn{1}{c}{\textbf{82.33}}&\multicolumn{1}{c}{-}&\multicolumn{1}{c}{-}&\multicolumn{1}{c|}{-}&\multicolumn{1}{c}{87.69}&\multicolumn{1}{c}{\textbf{89.20}}&\multicolumn{1}{c}{88.32}&\multicolumn{1}{c}{-}&\multicolumn{1}{c}{-}&\multicolumn{1}{c}{-} \\
	\hline
	\multicolumn{4}{c|}{\textbf{Ours}}&\multicolumn{3}{c|}{ConvNeXt-base}&\multicolumn{1}{c}{\textbf{88.11}}&\multicolumn{1}{c}{\textbf{91.51}}&\multicolumn{1}{c|}{\textbf{89.13}}&\multicolumn{1}{c}{\textbf{85.05}}&\multicolumn{1}{c}{84.11}&\multicolumn{1}{c}{81.12}&\multicolumn{1}{c}{\textbf{80.18}}&\multicolumn{1}{c}{\textbf{83.51}}&\multicolumn{1}{c|}{\textbf{80.36}}&\multicolumn{1}{c}{\textbf{88.12}}&\multicolumn{1}{c}{88.76}&\multicolumn{1}{c}{\textbf{88.54}}&\multicolumn{1}{c}{\textbf{78.54}}&\multicolumn{1}{c}{\textbf{76.34}}&\multicolumn{1}{c}{\textbf{75.91}} \\
	 \bottomrule
\end{tabular}}
\caption{Benchmark results in RAP, RAPzs, PETA, PETAzs and PA100k. Our method is compared with various notable SOTA methods. To make a fair comparison, for RAP, PETA and PA100k, we adpot the baseline results of Resnet-50 and ConvNeXt-base respectively from~\protect\cite{jia2021rethinking} and~\protect\cite{Specker2022UPARUP}, and the results reported in the original literature for each prior work. For RAPzs and PETAzs, we refer scores from the datasets work~\protect\cite{jia2021rethinking} with priority, next to use public code of the preivous work, if any, to reproduce the results. Since there is no baseline reported for ConvNeXt-base on RAPzs and PETAzs, corresponding results are based on our experiments. If no result is reported as a certain setting or the public code is not available for convincible testing on RAPzs and PETAzs, it is marked as $-$. *results are produced with additional data augmentations. All values are percentages and the highest scores are marked by bold fonts.} 
\label{benchmark}
\end{table*}

In a nutshell, attribute-disentangled learning is secured by our method from two sides. One, given the specific feature of all attributes, we render the posterior estimate of a certain attribute exclusively associated to the feature component of its own, tasking the rest specific features, on which other attributes are inferred, to preserve no clues informative about this given attribute. Second, serving as an additional virtue of Eq.\ref{train}, the biased attribute interdependencies are strongly dismissed from classifier learning owing to the scheme that the ground truth employed in Eq.\ref{train} is randomly generated for each attribute, and therefore can be of any possible pattern in term of attributes co-occurrence, making classifier no way to trace on the attribute correlations embedded in the limited dataset, and thus generalize better. Efficiently, both sides are achieved in a holistic, end-to-end manner.

\section{Experiments}

\paragraph{Data and Evaluation Metric.} For the benchmark datasets, PETA~\cite{2014Pedestrian}, along with the two largest public pedestrian attribute datasets RAP~\cite{li2016richly} and PA100k~\cite{liu2017hydraplus}, are adopted for evaluation. Detailed dataset information and usage are consistent to those in~\cite{jia2021rethinking}. We also test our methods on two realistic datasets of RAPzs and PETAzs stated and released in~\cite{jia2021rethinking}. As for the evaluation metrics, the label-based metric mean Accuracy (mA), which takes an average over all attributes' classification accuracy on the positive and negative samples, and two instance-based metrics Recall and F1-score (F1) are considered. we do not present Precision since it can be basically inferred when Recall and F1 are told.

\paragraph{Network and Training Details.} We adopt Resnet-50~\cite{2016Deep} and ConvNeXt-base~\cite{liu2022convnet} as backbones to study the efficacy and compatibility of our method under feature extractors of both classical and up-to-date designs. For the training details, Adam solver is applied without Nesterov momentum. The learning rate starts at 1e-4 and decays by a factor of 10 in a manner of multistep. If not specifically stated, the results in this section are produced by Eq.\ref{train} with efficiency. We refer readers to the code of this work in the supplementary material for further details.

\paragraph{Benchmark Results.}
Following the benchmark protocol provided in~\cite{jia2021rethinking}, our method is compared with recent notable SOTA approaches MsVAA~\cite{2018Deep}, MTMS~\cite{10.1145/3343031.3351003}, ALM~\cite{2019Improving}, MTA-Net~\cite{2020Pedestrian}, AR-BiFPN~\cite{2020Relation}, VAC~\cite{2020Visual}, JLAC~\cite{2020Relation}, SSC~\cite{Jia_2021_ICCV}, DAFL~\cite{DBLP:conf/aaai/JiaGHCH22} and Label2Label~\cite{2022Label2Label}. Since we start at a training setup that involves less data augmentation methods, the baseline performances on these datasets could be about 1\% mA inferior to some of the compared methods. To mitigate this gap, we also present experimental results of applying the data augmentation settings akin to JLAC (additional random scaling, rotation, translation, cropping, erasing and adding random gaussian blurs). The overall results are reported in Table.\ref{benchmark}. It highlights that across all settings, our prescription achieves performance at least comparable to others. Please note that our method is highly efficient since for test samples, the extra operations over the baseline models, are only one single FC layer applied to produce attribute-specific features, in stark contrast to the prior arts paying a premium in term of computational cost. Therefore, one potential of it is that, one could just make full use of the model parameters being largely saved by our lightweight framework, to plug a better-but-wider backbone, like ConvNeXt for further significant improvements.

Basically, one might notice that our proposal works on PA100k better than RAP and PETA. It is not surprising, since, as pointed out by~\cite{jia2021rethinking}, about 31.5\% and 57.7\% of pedestrian identities in the test set of RAP and PETA are identical to those in their respective training set. Thus, for PETA and RAP, memorizing pattern of the biased attributes co-occurrence in training set can be conducive for model's test-set performance, making existing approaches overestimated on them. For this, we also report in Table.\ref{benchmark} the experimental results on the PETAzs and RAPzs, which are respectively formed from PETA and RAP to follow the zero-shot setting of pedestrian identities, i.e., no overlapped identities between their training and test sets. Therefore, the results on PETAzs and RAPzs, along with the results on PA100k, are much convincible~\cite{Jia_2021_ICCV}, and thus should be attached of more practical implications, on which our method outperforms others with considerable margins. Also, it should be noticed that the concepts of existing works only fuel trivial boost on these realistic datasets, implying that previous development of PAR on PETA and RAP may partly come from better modeling of the common bias in training and test sets. Overall, our method excels in both the recognition performance and practical applicability of realistic PAR.

\paragraph{Ablation Studies.}
In this sub-section, we investigate the effectiveness and characteristics of each technical contribution. To better discern between attributes classifiers, we use PA100k and ResNet-50 in the following ablation studies.

\begin{figure}[t]
\centering
\includegraphics[height=3.8cm,width=8.35cm]{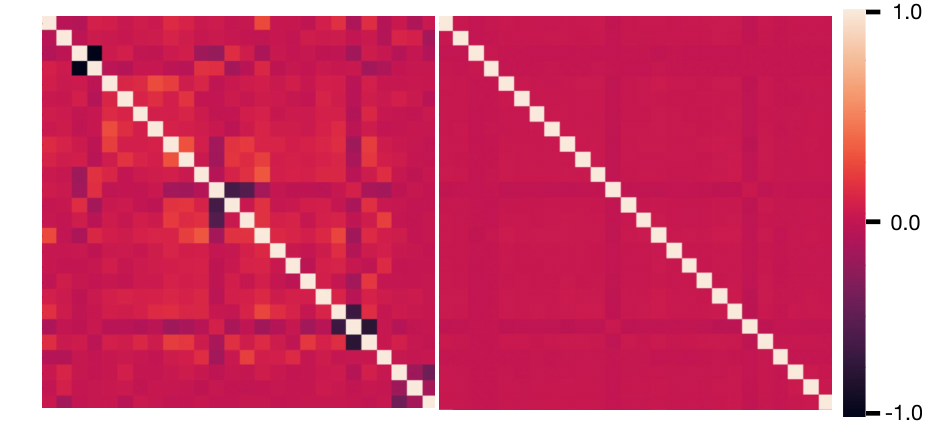}
\caption{The matrix of cosine similarities between PA100k attribute anchor features of Left: baseline model and Right: our method. The results clearly support that our work suppresses the undesirable modeling of biased attributes co-occurrence in classifier. Same to that in Figure.\ref{heatmap}, weights in the last FC layer of a trained Resnet-50 are employed as the attribute anchor features.}
\label{our}
\end{figure}

\paragraph{Posterior-invariant learning vs. MixUp-based efficient learning.} In this paper, we present two pipelines for attributes-disentangled feature learning, the one is directly derived from the goal of posterior-invariant learning to minimize the mutual information between one certain attribute and its irrelevant features specific to other attributes, as described in Eq.\ref{regular}. While effective as shown in the first row of Table.\ref{mixup}, it requires multiple inferences for a single sample, which hinders the training efficiency and increases the memory usage when the number of attributes is large. To address this issue, a second pipeline for efficient single-inference-every-instance is introduced in Eq.\ref{train}. It extends the concept behind MixUp to significantly lower the training cost, and achieves performance similar to Eq.\ref{regular}, as can be seen in the third row of Table.\ref{mixup}. During our multiple rounds of experiments, we find that Eq.\ref{train} can always yield slightly better mA, and we credit such improvement into that the biased attribute correlation within dataset is totally discarded during the training of classifier in Eq.\ref{train}, since the labels used in it are attribute-wise randomly generated thus can form any possible pattern of attribute co-occurrence, making classifier hardly to memorize the limited attribute co-occurrence presenting in the dataset, and thereby generalize better, as shown in Figure.\ref{our}.

\paragraph{Norm-direction separated interpolation vs. plain feature  interpolation.} We apply the proposed norm-direction separated interpolation in Eq.\ref{regular} and Eq.\ref{train} to fully exploit the possible variations of attribute-specific features, as graphed in Figure.\ref{holder}. The results in Table.\ref{mixup} clearly demonstrate that this trick, in tandem with both pipelines, outperforms the plain interpolation used in previous work and thus can be applied in other related fields of research like recognition with unstable image quality for enhanced robustness and performance.

\begin{table}[t]
\setlength{\abovecaptionskip}{10pt}
\setlength{\belowcaptionskip}{-0.cm}
\footnotesize
	\centering
         \renewcommand{\arraystretch}{1.0}
	\begin{tabular}{cc|cccc}
	 \toprule
	 \multicolumn{2}{c|}{\textbf{Our model}}&\multirow{2}{*}{\textbf{mA}}&\multirow{2}{*}{\textbf{Recall}}&\multirow{2}{*}{\textbf{F1}}\\
	\multicolumn{1}{c}{\textbf{Pipeline}}&\multicolumn{1}{c|}{\textbf{NDSI}}\\
	 \midrule
	 \multicolumn{1}{c}{Eq.\ref{regular}}&\multicolumn{1}{c|}{}&\multicolumn{1}{c}{83.20}&\multicolumn{1}{c}{89.27}&\multicolumn{1}{c}{86.96}\\
	 \multicolumn{1}{c}{Eq.\ref{regular}}&\multicolumn{1}{c|}{\Checkmark}&\multicolumn{1}{c}{83.79}&\multicolumn{1}{c}{89.51}&\multicolumn{1}{c}{87.06}\\
	\multicolumn{1}{c}{Eq.\ref{train}}&\multicolumn{1}{c|}{}&\multicolumn{1}{c}{83.83}&\multicolumn{1}{c}{89.04}&\multicolumn{1}{c}{86.85}\\
	\multicolumn{1}{c}{Eq.\ref{train}}&\multicolumn{1}{c|}{\Checkmark}&\multicolumn{1}{c}{84.53}&\multicolumn{1}{c}{89.13}&\multicolumn{1}{c}{87.01}\\
	\midrule
	\multicolumn{2}{c|}{\multirow{1}{*}{\textbf{Method}}}&\textbf{mA}&\textbf{Recall}&\textbf{F1}\\
	 \midrule
	 \multicolumn{2}{c|}{Baseline}&\multicolumn{1}{c}{80.38}&\multicolumn{1}{c}{87.01}&\multicolumn{1}{c}{87.05}\\
	 \multicolumn{2}{c|}{MixUp}&\multicolumn{1}{c}{79.73}&\multicolumn{1}{c}{85.22}&\multicolumn{1}{c}{86.75}\\
	 \bottomrule
	\end{tabular}
	\caption{The breakdown effect for each technical component of the introduced method, in which NDSI is short for the norm-direction separated interpolation depicted in Figure.\ref{holder}. Also, the comparison of our method against MixUp. }
	\label{mixup}
\end{table}

\paragraph{Mixup vs. Proposed Methods.} Eq.\ref{train} directly optimizes over the interpolated features with their correspondingly interpolated labels, in a similar way as that of Mixup. Therefore, it might be questioned that our method's superiority can fundamentally come from tailoring on MixUp. However, as shown in Table.\ref{mixup}, sample interpolation is not the case here for bringing significant performance boost, since Mixup delivers even negative increase in mA. Moreover, our pipeline of Eq.\ref{regular}, which does not optimize over the augmented samples, can score comparatively with Eq.\ref{train}. It implies that, what essentially works for Eq.\ref{train} is the strategy that we use feature interpolation to enable a random and differentiated transform over each attribute's specific feature, and empower model with the consequential mutual information minimization learning by Eq.\ref{regular} or Eq.\ref{train}. On a higher level, data augmentations are methods creating data points beyond the training set to reduce the various bias embedded in dataset. Under such perspective, it brings no bad to comprehend our method as a data augmentation trick that augments the pattern of attributes co-occurrence, out of the less representative dataset. 

\begin{figure*}[t]
\centering
\includegraphics[height=3.86cm,width=17.82cm]{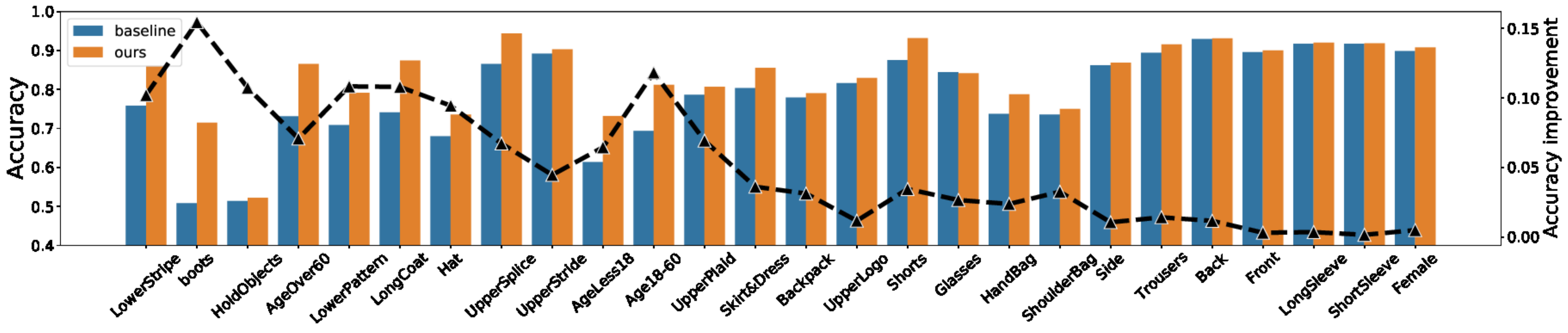}
\caption{Accuracy results comparison of our method with the baseline work on all attributes of PA100k. From left to right, the attributes are arranged in a decreasing order of the distance between their respective positive ratio and 0.5. The black dashed line marks the variation of accuracy increase on each attribute, which is mean smoothed with the windows size of 2 for better visualization.}
\label{improve}
\end{figure*}

\paragraph{Analysis on Improvements.}
In Figure.\ref{improve}, we draw the per-attribute accuracy results on PA100k dataset of the proposal and the baseline work. We find that our work is capable of sizably raising scores on almost all attributes. It might serve as a relief to the likely concern that discarding totally the use of attributes correlation could be detrimental to the recognition of some certain type of attributes, for which a further study is presented in the next parts. Generally, it can be seen that improvements are considerable on attributes with positive ratio closer to 1 or 0. It is expectable as attributes with inadequate chances to appear or disappear are likely to be limited in the captured or non-captured scenes and identities in term of the attributes interdependency. By our framework, on one side the inference dependence of an imbalanced attribute to its frequently or hardly co-occurred attributes is greatly undermined, reducing the bias within feature learning. On another side, aforementioned for Figure.\ref{our}, the memorization effect of attributes correlation in classifier is also repressed. Thanks to these two factors, the recognition of attributes with most potential to be correlated can be facilitated utterly, whilst for attributes appearing evenly, the improvements can be trivial. It also accounts for the reason that our method works better on label-based metric than instance-based metric, since the growth of mA is actually bottlenecked by the imbalanced attributes, while instance-based metrics are relatively not.

\paragraph{On the Attribute Interdependency Modeling.}
Intuitively, for attributes not causally independent to each other, discovering and utilizing the relations among them might be conducive for robust information exchanges and propagation in PAR. For this, turning back to seek help from previous work that precisely models the attributes interdependency seems suboptimal, since without elaborate expert knowledge use, there is no feasibilities of learning, for these methods to discern the welcomed attributes interdependency from the intricate attributes co-occurrence bias, and exclusively dismiss use of the latter. Noting in Figure.\ref{improve}, the baseline model is error-prone not only for attributes appearing barely, but those occurring often like 'Age18-60' (positive ratio above 0.9) so, such an indiscriminative modeling of all attributes correlations is overall detrimental to most imbalanced attributes, making PAR strongly bottlenecked.

Moreover, there is another drawback of learning with attribute interdependencies - even the causally robust ones like the mutual exclusiveness among ages groups ( 'age $\textless$ 18', '18 $\le$ age $\le$ 60' and 'age $\textgreater$ 60' in PA100k are of a multi-class relation rather than that of multi-label), for which we call the infestation of attribute independency. As shown in Figure.\ref{our}, two attributes tending to (not to) co-occur with each other would also prefer to (not to) co-occur with other attributes, representing as the dark or light stripes spanning across the anchor feature similarity matrix. Whereas these traversing lines do not show up in the Pearson correlation coefficient matrix of labels, it is actually by-produced during learning of the attributes correlations. This inclination of hallucinating new interdependency bias from the dataset attributes interdependency is smoothed out by ours, greatly yet not totally, as there are still lines can be eyed in the heat map.

\begin{table}[t]
\setlength{\abovecaptionskip}{10pt}
\setlength{\belowcaptionskip}{-0.cm}
\footnotesize
	\centering
         \renewcommand{\arraystretch}{1.0}
	\begin{tabular}{lcccc}
	\toprule
	 \textbf{Age attributes}&\textbf{Baseline}&\textbf{Ours}&\textbf{Robust-A}&\textbf{Robust-B}\\
	 \midrule
	 AgeLess18&61.36&75.67&75.72&75.67\\
	 Age18-60&69.36&82.21&81.95&82.22\\
	 AgeOver60&73.15&84.87&84.94&84.87\\
	 \bottomrule
	\end{tabular}
	\caption{On three causally related age attributes, we compare the accuracy results of baseline, our work and the described robust PARs. For Robust-A, we use weighted loss to soften the class imbalance.}
	\label{robust}
\end{table}

\paragraph{Towards Robust PAR.}
Ideally, a PAR of robustness should be capable of inferring on a-prior concrete attributes interdependencies while disregarding the others. Solutions of ease towards this aim can stand on the base of attributes-disentangled feature learning. Taking the age groups in PA100k as example, on a trained model of attributes-disentangled learning, one could train an additional classifier head of multi-class only for age prediction (namely, Robust-A), or conduct certain back-end processings onto the outputs to rectify or unify the age predictions (Robust-B). Here, we realize both in Table.\ref{robust}. For Robust-B, we output only the age attribute of highest confidence score. Compared to our method, these robust PARs yield negligible accuracy increase in test set. However, we still emphasize on the importance of such designs since they secure the reliability of PAR, and might make difference in realistic conditions. 

\section{Conclusion}
We present a novel method of attribute-disentangled feature learning to enhance the robustness of PAR. Without any sophisticated or time-consuming frameworks, leveraging a simple information-theoretic regularizer, it is ensured that the pedestrian attributes are inferred without considering the biased attributes interdependency inherent to dataset, in order to enable an attribute recognition mechanism respecting us human. The comprehensive experiments demonstrate that our proposal reaches SOTA performance with appealing merits like better generalizability and applicability in realistic scenarios. Importantly, our theorized perspectives of attribute disentanglement learning differs from the paradigms of previous methods, even advancing in an opposite direction, heuristically exploring a promising avenue for future work. 

\section*{Acknowledgments}
This work was partially supported by the "Pioneer" and "Leading Goose" R\&D Program of Zhejiang (Grant No. 2023C01030), the National Natural Science Foundation of China (No.62122011, U21A20514), and the Fundamental Research Funds for the Central Universities. (Corresponding Author: Hai-Miao Hu)

\bibliographystyle{named}
\bibliography{ijcai23}

\end{document}